\documentclass[sigconf]{acmart}
\usepackage{multirow,color}
\usepackage{subcaption, tabularx}

\AtBeginDocument{%
  \providecommand\BibTeX{{%
    \normalfont B\kern-0.5em{\scshape i\kern-0.25em b}\kern-0.8em\TeX}}}

\copyrightyear{2023}
\acmYear{2023}
\setcopyright{acmlicensed}\acmConference[MMAsia '23]{ACM Multimedia Asia 2023}{December 6--8, 2023}{Tainan, Taiwan}
\acmBooktitle{ACM Multimedia Asia 2023 (MMAsia '23), December 6--8, 2023, Tainan, Taiwan}
\acmPrice{15.00}
\acmDOI{10.1145/3595916.3626459}
\acmISBN{979-8-4007-0205-1/23/12}

\begin{document}

\title{Cross-Modal Retrieval for Motion and Text via DropTriple Loss}


\author{Sheng Yan$^{1}$, Yang Liu$^{2}$, Haoqiang Wang$^{1}$, Xin Du$^{1}$, Mengyuan Liu$^{3\dag}$, Hong Liu$^{3}$}
\affiliation{%
  \institution{{$^{1}$} School of Artificial Intelligence, Chongqing University of Technology, China}
  \country{}
}
\affiliation{%
  \institution{{$^{2}$} College of Computer Science, Sichuan University, China}
  \country{}
}
\affiliation{%
  \institution{{$^{3}$} Key Laboratory of Machine Perception, Shenzhen Graduate School, Peking University, China}
  \country{}
}
\affiliation{
    \institution{\small{$^\dag$Corresponding author (e-mail: nkliuyifang@gmail.com)}}
    \country{}
}

\renewcommand{\shortauthors}{Sheng Yan, Yang Liu and Haoqiang Wang, et al.}



\begin{abstract}
Cross-modal retrieval of image-text and video-text is a prominent research area in computer vision and natural language processing. However, there has been insufficient attention given to cross-modal retrieval between human motion and text, despite its wide-ranging applicability. To address this gap, we utilize a concise yet effective dual-unimodal transformer encoder for tackling this task. Recognizing that overlapping atomic actions in different human motion sequences can lead to semantic conflicts between samples, we explore a novel triplet loss function called DropTriple Loss. This loss function discards false negative samples from the negative sample set and focuses on mining remaining genuinely hard negative samples for triplet training, thereby reducing violations they cause. We evaluate our model and approach on the HumanML3D and KIT Motion-Language datasets. On the latest HumanML3D dataset, we achieve a recall of 62.9\% for motion retrieval and 71.5\% for text retrieval (both based on R@10). The source code for our approach is publicly available at https://github.com/eanson023/rehamot.
\end{abstract}





\maketitle

\section{Introduction} \label{sec:intro}

Recently, there has been significant interest in the integration of natural language and images \cite{karpathy2015deep,li2017identity,radford2021learning, ramesh2022hierarchical, li2021align}. Cross-modal text-image retrieval has become a prominent research area \cite{faghri2017vse++, zhao2023generative, wang2023quaternion}. However, the retrieval problem that connects 3D human motion has yet to be extensively explored on a large scale. The ability to automatically match natural language descriptions with accurate 3D human motion (i.e., 3D human pose sequences) will open the door to numerous applications. For instance, video surveillance and security applications can utilize language descriptions and human motion to search for and identify specific events and behaviors.

\begin{figure}[t]
\centering
\includegraphics[width=0.95\columnwidth]{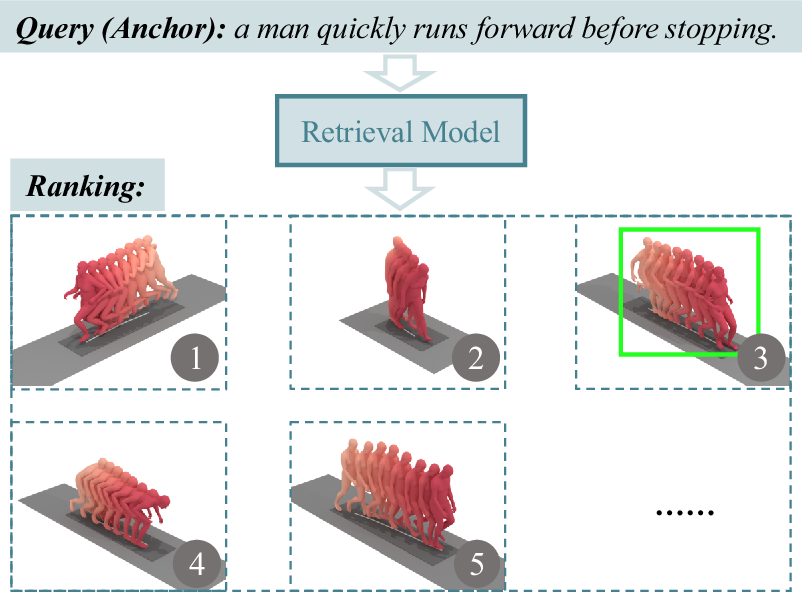}
\vspace{-10pt}
\caption{As an example of motion retrieval:  Given a textual query (anchor), the retrieval model searches for positive motion sample (green box) in the motion library. Likewise, text retrieval follows a similar procedure.}
\label{fig:fig1}
\vspace{-15pt}
\end{figure}

This study focuses on achieving cross-modal retrieval between 3D human motion and text (as illustrated in Figure \ref{fig:fig1}). Currently, research on motion synthesis has forged a bridge between human motion and natural language. TEMOS \cite{petrovich2022temos} introduces the Variational Autoencoder (VAE) architecture into this task, allowing the generation of diverse motion sequences based on a text description. MDM \cite{tevet2022human} incorporates the Diffusion Model into this task to generate natural and expressive human motions. Relevant to our work, Delmas et al. \cite{delmas2022posescript} utilize rich natural language and 3D human poses for bidirectional retrieval, providing a detailed pose annotation pipeline. However, their research is limited to static geometric human poses, which can be understood as a 3D still image, essentially falling within the scope of text-image retrieval.

Compared to retrieving static human poses, human motion sequences contain more information and higher dimensions. Establishing an effective temporal modeling model to learn embeddings of human motion and text descriptions is a key challenge. We propose a concise yet effective dual-unimodal encoder to encode, aggregate, and project features of motion and text sequences into a joint embedding space. The  dual-unimodal encoder utilizes the attention mechanism \cite{vaswani2017attention} to interact and integrate information from different positions in the sequences, effectively capturing the long-term dependencies of the sequences.

In the joint space, a common approach is to employ contrastive learning to learn the similarity of motion-text pairs. This approach is based on defining positive (\textbf{Pos}) and negative samples (\textbf{Neg}s) with respect to \textbf{Anchor} and uses distance metric methods to associate embeddings from different modalities within the same space. As such, a flexible principle is established: pulling Anchor and Pos together in the joint space while pushing Anchor away from multiple Negs. This principle can be implemented in various ways, including max-margin loss \cite{hadsell2006dimensionality, tu2023dtcm}, triplet loss \cite{wang2014learning, wang2016learning}, and InfoNCE \cite{zhou2021self, yang2022supervised}. Notably, the triplet loss based on hard negative (\textbf{hard-Neg}) mining, known as the Max of Hinges Loss (MH Loss), achieved a significant breakthrough \cite{faghri2017vse++, liu2022regularizing, chen2022intra}. We extend this idea in our model. Unlike images, human motion can be understood as a combination of different atomic actions at multiple time steps, and there often exist overlapping atomic actions between different motions. Therefore, hard-Negs often have strong semantic associations, similar to Pos, and can even better reflect the content of the Anchor. As evidence in Figure \ref{fig:fig1}, the hard-Neg (\textsf{Ranking 1}), similar to the Pos (\textsf{Ranking 3}), both indicating the "run forward" motion described by the Anchor. We refer to such hard-Negs as false negatives (\textbf{false-Neg}s). Conventional MH Loss, by mining these false-Negs, may separate samples that are actually strongly correlated, which is overly harsh and unreasonable.

 
The above-mentioned problem is another key challenge in this task. By sorting the intra-modal similarity within the Neg set, we attempt to determine a reasonable threshold that represents the boundary of semantic similarity. Negs exceeding this threshold are considered false-Negs. We discard false-Negs from the Neg set and focus on mining remaining hard-Negs for triplet training, thereby reducing violations they cause. We refer to this triplet loss based on the discard mining approach as DropTriple Loss. By comparing it with MH Loss on the latest HumanML3D \cite{guo2022generating} and KIT Motion-Language \cite{plappert2016kit} datasets, we validate the effectiveness of this mining approach.

\textbf{Overall, our main contributions are as follows:} (\textbf{i}) We investigate the overlooked task of cross-modal retrieval between motion and text by constructing a concise yet effective model. (\textbf{ii}) We propose the DropTriple Loss, which addresses unnecessary semantic conflicts caused by false negative samples.

\section{Method} \label{sec:method}

\begin{figure*}[h]
\centering
\includegraphics[width=0.95\textwidth]{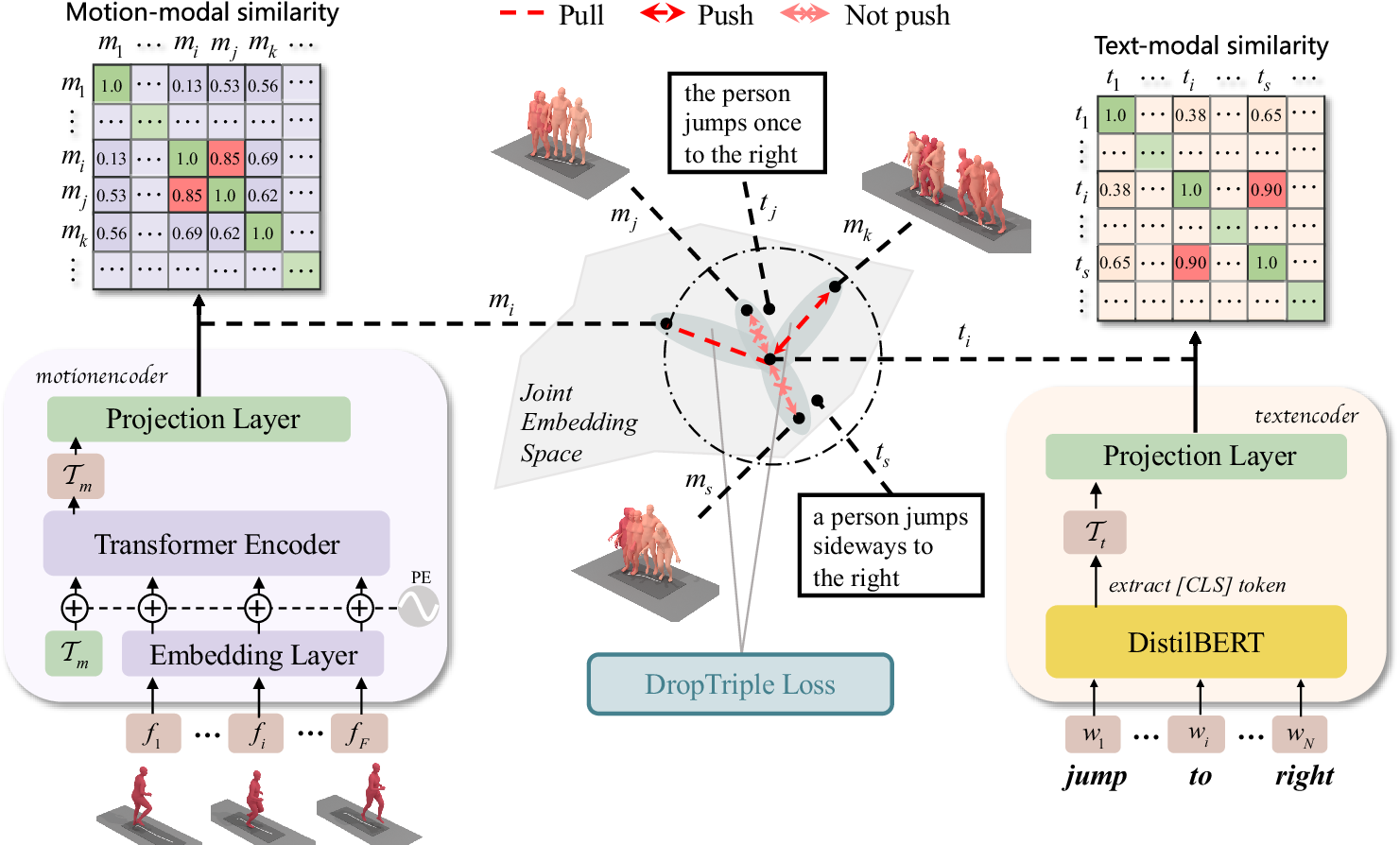}
\vspace{-5pt}
\caption{Our proposed framework encodes and aggregates the motion and text inputs separately in their respective encoders. Finally, the outputs are mapped to the joint embedding space through a projection layer. Within the same training batch, the DropTriple Loss discards mining false-Negs $m_{j}$ and $m_{s}$, while pushing the genuinely hard-Neg $m_{k}$ away.}
\label{fig:model_design}
\vspace{-10pt}
\end{figure*}

In this section, we introduce the motion-text cross-modal retrieval task (Section \ref{sec_task_define}), the model architecture used (Section \ref{sec_model_architecture}), and the objective function (Section \ref{sec_learning_objective}), which includes the standard triplet loss, the definition of false negatives, and our custom DropTriple Loss for motion-text retrieval.

\subsection{Task Definition} \label{sec_task_define}

\textbf{Text descriptions} represent the description of human motion in written natural language sentences, such as in English. The sentence contains an accurate sequence of actions, such as \textit{"a person holds their arms out, lowers them, then walks forward and sits down"}. The data structure is a word sequence $\boldsymbol{t} = ({w_1},\ldots,{w_N}),\text{ } w \in {\mathbb{R}}^{D_w}$ of length $N$ (with each word counted as 1) from the English vocabulary, where $D_w$ represents the word embedding dimension.

\textbf{3D human motion} is defined as a series of human poses $\boldsymbol{m}=({{f}_{1}},\ldots,{{f}_{F}}),\text{ } {f} \in {\mathbb{R}}^{D_p}$, where $F$ represents the number of time frames. Each posture ${f}$ corresponds to the representation of an articulated human body. In this paper, we use joint rotations, joint positions, and other related information to represent the body motion of each posture $f$, forming a $D_{p}$-dimensional feature vector. A more detailed definition will be given in Section \ref{ssec:pose_representation}.

\textbf{Task objective.} For motion retrieval, given a query in text, the task is to retrieve the most relevant human motion sequences from a database. Similarly, for text retrieval, the query is a human motion, and the task is to retrieve relevant text. The objective is to maximize recall at $K$ ($R@K$), where the fraction of queries ranked among the top $K$ items returned is the most relevant \cite{hodosh2013framing} ($K$ is typically 1-10). Let $\mathcal{D}=\{({{m}_{i}},{{t}_{i}})\}_{i=1}^{I}$ be the training set of motion-text pairs. We call $({{m}_{i}},{{t}_{i}})$ a positive pair, and $({{m}_{i}},{{t}_{j\ne i}})$ a negative pair. Thus, we have $I$ positive pairs and ${{I}^{2}}-I$ negative pairs in the training set. To achieve satisfactory performance at $R@K$, we need to maximize the similarity of $I$ positive pairs in the training set, while minimizing the similarity of ${{I}^{2}}-I$ negative pairs.

\subsection{Model Architecture} \label{sec_model_architecture}

As illustrated in Figure \ref{fig:model_design}, influenced by TEMOS and various image-text models \cite{radford2021learning, he2020momentum, li2021align, liu2023temporal, zolfaghari2021crossclr}, we adopt a dual-branch unimodal network to extract motion and text embeddings and project them into a joint embedding space. For the motion branch, the encoder takes arbitrary-length pose sequences as input. Before feeding each body pose into the Transformer Encoder (\textbf{TMR Enc}), it is first embedded into a $D_{\ell}$-dimensional space in the embedding layer. Since we embed arbitrary-length sequences into one space (sequence-level embedding), we need to aggregate the time dimension. To achieve this, a learnable token $\mathcal{T}_{m}$ is appended to the embedded pose sequence as a temporal aggregator. The resulting input to TMR Enc is the sum of positional encoding, given in the form of a sine function. By extracting the first output of TMR Enc that corresponds to the token (discarding the rest), we obtain the motion feature $\mathcal{T}_{m}$. For the text branch, we employ the pre-trained expert model DistilBERT \cite{sanh2019distilbert} as the backbone network and take its $D_{w}$-dimensional [CLS] token $\mathcal{T}_{t}$ as the text feature. Unless otherwise specified, the weights of DistilBERT are frozen. The aforementioned networks can be parameterized as $\mathscr{M}_{enc}(\cdot;\theta_{\phi})$ and $\mathscr{T}_{enc}(\cdot;\theta_{\psi })$ for the motion and text branches, respectively.

Next, we use the projection layers $h(\cdot;W_{h})$ and $g(\cdot;W_{g})$ to define the embeddings mapped to the joint embedding space.  We also define a similarity function $\mathcal{S}(\cdot,\cdot)$ to measure the similarity between them. Mathematically, the entire process can be formulated as:
\begin{equation}
\mathcal{S}(m, t)=h\left(\mathcal{T}_{m} ; W_{h}\right) \cdot g\left(\mathcal{T}_{t}; W_{g}\right) \label{eq:model}
\end{equation}
where $\cdot$ denotes inner product, $W_{h} \in \mathbb{R}^{{D_{\ell}} \times D}$ and $W_{g} \in \mathbb{R}^{{D_{w}} \times D}$. The relevant features $\mathcal{T}$ are represented by selecting the first vector from the output sequence of $\mathscr{M}_{enc}(m;\theta_{\phi})$ or $\mathscr{T}_{enc}(t;\theta_{\psi})$, corresponding to the respective token. Before computing the inner product, we apply $\ell_{2}$-normalization to the embeddings. In this case, the inner product is equivalent to cosine similarity. Let $\theta = \{ W_{f},W_{g}, \theta_{\phi} \}$ be the overall model parameters, and if we also need to fine-tune the $\mathscr{T}_{enc}$ network, then $\theta_{\psi}$ will be included in $\theta$ as well.

\subsection{Learning Objective} \label{sec_learning_objective}


\subsubsection{SH Loss \& MH Loss} \label{SH_Loss_MH_Loss}

Using the standard triplet loss, SH Loss (Sum of Hinges Loss), can achieve the aforementioned task objective (Section \ref{sec_task_define}), and it has been widely applied in other cross-modal retrieval tasks \cite{kiros2014unifying, zhao2023generative}. SH Loss aims to learn the model parameters $\theta$ by minimizing the cumulative loss over the training data $\mathcal{D}=\{({m}_{i},{t}_{i})\}_{i=1}^{I}$, given by:
\begin{equation}
\begin{split}
\mathcal{L}_{SH}(\theta,\mathcal{D})=\sum\limits_{i=1}^{I}{\sum_{\hat{t} \in \mathcal{Q}_{\text{T}} }[\alpha -\mathcal{S}(m_{i},t_{i})+\mathcal{S}(m_{i},\hat{t}) ]_{+}}\\
+\sum\limits_{i=1}^{I}{\sum_{\hat{m} \in \mathcal{Q}_{\text{M}} }[\alpha -\mathcal{S}(m_{i},t_{i})+\mathcal{S}(\hat{m},t_{i})   ]_{+}}   \label{eq:sh_loss}
\end{split}
\end{equation}
where $\alpha$ is a margin hyperparameter, and ${{\left[ x \right]}_{+}}\equiv max(0,x)$. $\mathcal{Q}_{\text{M}} = \{ m_j \mid j \in [I] \setminus \{i \} \}$ and $\mathcal{Q}_{\text{T}} = \{ t_j \mid j \in [I] \setminus \{i \} \}$ represent the sets of Negs for motion and text, respectively. $\mathcal{S}(\cdot,\cdot)$ refers to the similarity measurement function mentioned in Eq. \ref{eq:model}

Faghri et al. \cite{faghri2017vse++} demonstrated that the SH Loss (Eq. \ref{eq:sh_loss}) can lead to local minima when multiple negatives with small violations dominate the loss. To tackle this issue, they proposed the Max of Hinges (MH) Loss, which focuses on the hardest Neg to mitigate this problem:
\begin{equation}
\begin{split}
\mathcal{L}_{MH}(\theta,\mathcal{D})=\sum\limits_{i=1}^{I}{[\alpha -\mathcal{S}(m_{i},t_{i})+\mathcal{S}(m_{i},{t}') ]_{+}}\\
+\sum\limits_{i=1}^{I}{[\alpha -\mathcal{S}(m_{i},t_{i})+\mathcal{S}({m}',t_{i})]_{+}} \label{eq:mh_loss}
\end{split}
\end{equation}
where ${t}'=argmax_{t_{j} \in \mathcal{Q}_{\text{T} }} \mathcal{S}(m_{i}, t_{j})$ and ${m}'=argmax_{m_{j} \in \mathcal{Q}_{\text{M} }} \mathcal{S}(m_{j},t_{i})$ denote the hardest Negs from their respective Neg sets. Recent studies \cite{liu2022regularizing, zhao2023generative} have shown that the MH Loss performs better than the SH Loss.

\subsubsection{False Negative Sample Definition} \label{FNS_define}


False negative sample (\textbf{false-Neg}) actually have strong semantic overlap with Pos. For example, in Figure \ref{fig:fig1}, Ranking 1 (false-Neg) and Ranking 3 (Pos) illustrate such cases. By contrasting these unwanted false-Neg pairs, the network is encouraged to discard their common features in the learned embeddings, which goes against the common assumption in contrastive learning that having enough Negs helps to learn better embeddings \cite{he2020momentum, yuan2021multimodal,long2022multi}, because the model contrasts more semantic embeddings in each training batch. Therefore, when the number of false-Negs is large, frequent semantic conflicts \cite{zolfaghari2021crossclr} can hinder the algorithm from learning good embeddings.

As mentioned in the introduction (Section \ref{sec:intro}), we need to identify false-Negs that are equivalent to Pos from the Neg set $\mathbf{Q}_{\text{T/M}}$. For simplicity, we will describe the definition of the false-Neg set $\mathcal{Y}_{\text{M}}$ using motion retrieval as an example. As shown in Figure \ref{fig:model_design}, given an Anchor text $t_{i}$ with its relevant Pos motion $m_{i}$, we consider Neg $m_{j} \in \mathcal{Q}_{\text{M}}$ with high similarity to the Pos $m_{i}$ as false-Neg, based on the computed similarity in the motion modality (Figure \ref{fig:model_design}, top-left). A threshold $\delta$ is used to control the level of similarity required for a Neg to be defined as a false-Neg. In this case, $\delta$ can be set to 0.7. Mathematically, the set $\mathcal{Y}_{\text{M}}$ containing false-Negs can be written as follows:
\begin{equation}
\begin{split}
\mathcal{Y}_{\text{M}}=\{ m_{j} \mid \mathcal{S}(m_{i},m_{j})> \delta ,\forall m_{j} \in \mathcal{Q}_{\text{M}} \}
\end{split}
\end{equation}
the threshold $\delta$ represents the boundary of similarity between Negs and Pos, and samples in the set $\mathcal{Y}_{\text{M}}$ can be considered as entirely false-Negs. Moreover, we argue that solely considering the similarity in the motion modality is insufficient to identify all false-Negs. This approach should be extended to the text modality as well. As shown in Figure \ref{fig:model_design} (top-right), if there exists a text $t_{s}$ that exhibits high similarity with the Anchor text $t_{i}$, then the relevant motion $m_{s}$ for text $t_{s}$ in the training set $\mathcal{D}$ should also be included in the false-Neg set $\mathcal{Y}_{\text{M}}$:
\begin{equation}
\begin{split}
\mathcal{Y}_{\text{M}}=\{ m_{j} \mid \mathcal{S}(m_{i},m_{j})> \delta _{hetero} ,\forall m_{j} \in \mathcal{Q}_{\text{M}} \} \\
\cup \{ \mathcal{F}(t_{s}) \mid \mathcal{S}(t_{i},t_{s})> \delta _{homo} ,\forall t_{s} \in \mathcal{Q}_{\text{T}} \}
\end{split}
\end{equation}
where the hyperparameters $\delta_{hetero}$ and $\delta_{homo}$ denote represent the thresholds for the two modalities. We employ separate thresholds to control each modality. The function $\mathcal{F}(t)$ retrieves the relevant $m$ from the training set $\mathcal{D}$ based on $t$.

\subsubsection{DropTriple Loss} \label{sec_DropTriple_loss}

To reduce the impact of false-Negs in contrastive learning, we decided to remove all false-Negs (in each modality) from the Neg set. Therefore, we redefine the Neg sets for text retrieval and motion retrieval as: $\hat{\mathbf{N}}_{\text{T}}=\left\{t_{k} \mid \forall t_{k} \in \mathcal{Q}_{\text{T} }, t_{k} \notin \mathcal{Y}_{\text{T}}\right\}$ and $\hat{\mathbf{N}}_{\text{M}}=\left\{m_{k} \mid \forall m_{k} \in \mathcal{Q}_{\text{M} }, m_{k} \notin \mathcal{Y}_{\text{M}}\right\}$. By pruning the Neg sets in this manner, we can easily focus on mining genuinely hard-Negs for model training. The objective of the model is then formulated as the following:
\begin{equation}
\begin{split}
\mathcal{L}_{Drop}(\theta,\mathcal{D})=\sum\limits_{i=1}^{I}{[\alpha -\mathcal{S}(m_{i},t_{i})+\mathcal{S}(m_{i},\mathbf{{t}''}) ]_{+}}\\
+\sum\limits_{i=1}^{I}{[\alpha -\mathcal{S}(m_{i},t_{i})+\mathcal{S}(\mathbf{{m}''},t_{i})]_{+}} \label{eq:mild_loss}
\end{split}
\end{equation}
where $\mathbf{{t}''}=argmax_{\underline{t_{j} \in \hat{\mathbf{N}}_{\text{T}}}} S(m_{i}, t_{j})$ and $\mathbf{{m}''}=argmax_{\underline{m_{j} \in \hat{\mathbf{N}}_{\text{M}}}} S(m_{j},t_{i})$ respectively denote the hardest Negs from the pruned Neg sets. Optimizing these samples can further enhance performance.

In practice, for computational efficiency, we restrict the search for Negs to each mini-batch rather than the entire training set. Additionally, direct training of the model using either the MH Loss (Eq. \ref{eq:mh_loss}) or the DropTriple Loss (Eq. \ref{eq:mild_loss}) in motion-text retrieval tasks results in slow convergence. To address this, we adopt a curriculum learning \cite{bengio2009curriculum} strategy mentioned in \cite{faghri2017vse++}: before using these two losses, we \textbf{warm-up} the entire model for $\rho$ epochs using the SH Loss (Eq. \ref{eq:sh_loss}) to expedite training. We analyze this issue in the subsequent experimental section (Section \ref{sec:ablation_study}).

\begin{table*}[t]
\centering
\caption{Results of ablation experiment on HumanML3D and KIT-ML. Symbol (f) indicates fine-tuning the language model.}
\label{tab:reasult}
\vspace{-10pt}
\resizebox{\textwidth}{!}{%
\begin{tabular}{l|ccccccccc|ccccccccc}
\hline
\multirow{2}{*}{} & \multicolumn{9}{c|}{HumanML3D} & \multicolumn{9}{c}{KIT-ML} \\
 & \multicolumn{4}{c}{Motion Retrieval} & \multicolumn{4}{c}{Text Retrieval} &  & \multicolumn{4}{c}{Motion Retrieval} & \multicolumn{4}{c}{Text Retrieval} &  \\ \hline
Method & R@1$\uparrow$ & R@5$\uparrow$ & R@10$\uparrow$ & Med R$\downarrow$ & R@1$\uparrow$ & R@5$\uparrow$ & R@10$\uparrow$ & Med R$\downarrow$ & R-sum$\uparrow$ & R@1$\uparrow$ & R@5$\uparrow$ & R@10$\uparrow$ & Med R$\downarrow$ & R@1$\uparrow$ & R@5$\uparrow$ & R@10$\uparrow$ & Med R$\downarrow$ & R-sum$\uparrow$ \\ \hline
SH Loss \cite{kiros2014unifying} & 10.9 & 34.4 & 48.1 & 11.0 & 13.2 & 41.5 & 56.5 & 8.0 & 204.6 & 9.1 & 33.4 & 50.3 & 10.0 & 9.4 & 33.1 & 48.0 & 12.0 & 183.0 \\
MH Loss \cite{faghri2017vse++} & 10.7 & 35.1 & 48.9 & 11.0 & 14.0 & 43.3 & 58.1 & 7.0 & 210.1 & 10.6 & 33.9 & 50.7 & 10.0 & 11.2 & 32.3 & 43.4 & 15.0 & 182.1 \\
Our DropTriple Loss & 13.3 & 38.9 & 52.9 & 9.0 & 16.1 & 45.9 & 60.5 & 6.0 & 227.7 & 9.7 & 34.5 & 52.4 & 10.0 & 11.0 & 30.7 & 47.8 & 12.0 & 186.1 \\ \hline
\textbf{Our DropTriple Loss (f)} & \textbf{17.3} & \textbf{48.9} & \textbf{62.9} & \textbf{6.0} & \textbf{21.1} & \textbf{54.7} & \textbf{71.5} & \textbf{5.0} & \textbf{276.4} & \textbf{12.2} & \textbf{41.7} & \textbf{59.1} & \textbf{8.0} & \textbf{13.9} & \textbf{41.0} & \textbf{55.0} & \textbf{8.0} & \textbf{222.8} \\ \hline
\end{tabular}%
}
\end{table*}

\begin{table*}[htbp]
\centering
\caption{Motion retrieval results on HumanML3D. We visualize top 10 results for a given query. The images marked with a green box represent the ground-truth corresponding to the query.}
\label{tab:motion_retrieval}
\vspace{-10pt}
\begin{tabular}{@{}p{0.23\textwidth}p{0.7\textwidth}>{\raggedright\arraybackslash}c@{}}
\toprule
\multicolumn{1}{c}{\textbf{Query}} & \multicolumn{1}{c}{\textbf{Top-10 Retrieved Motions}} &  \\ \midrule

  \multirow{2}{0.23\textwidth}{\textit{someone working out as the get up off the floor.}} & \includegraphics[width=0.092\linewidth]{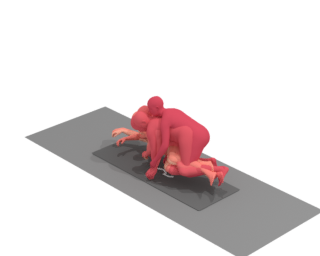}
\includegraphics[width=0.092\linewidth]{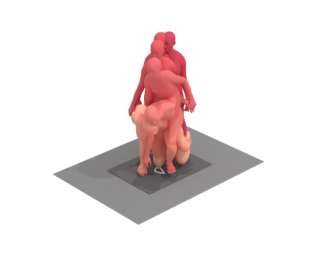}
\includegraphics[width=0.092\linewidth]{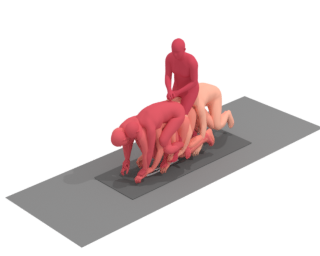}
\includegraphics[width=0.092\linewidth]{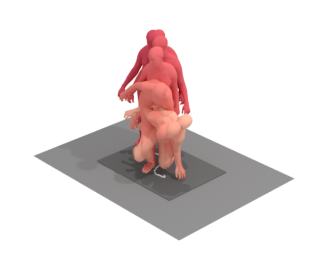}
\includegraphics[width=0.092\linewidth]{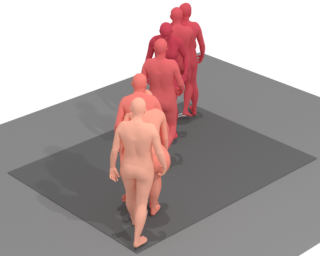}
\textcolor{green}{\fbox{\includegraphics[width=0.092\linewidth]{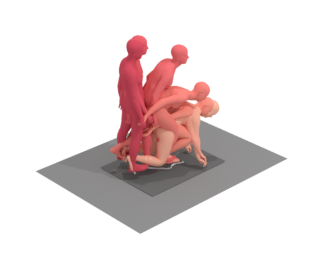}}}
\includegraphics[width=0.092\linewidth]{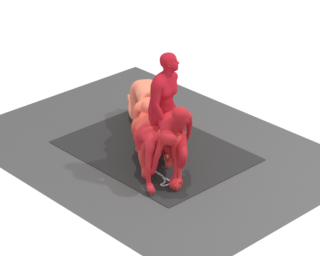}
\includegraphics[width=0.092\linewidth]{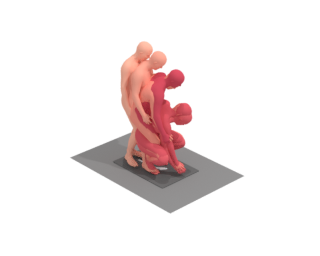}
\includegraphics[width=0.092\linewidth]{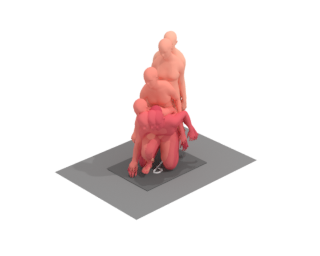}
\includegraphics[width=0.092\linewidth]{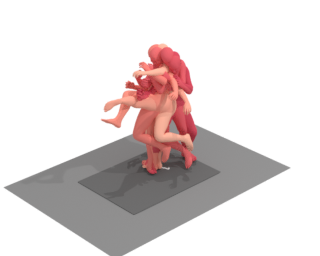} &  \raisebox{2.5ex}{\color[HTML]{C8414B} $\diamondsuit$} \\
 & \includegraphics[width=0.092\linewidth]{figs/results/t2m_000284/007080.png}
 \includegraphics[width=0.092\linewidth]{figs/results/t2m_000284/007746.png}
 \includegraphics[width=0.092\linewidth]{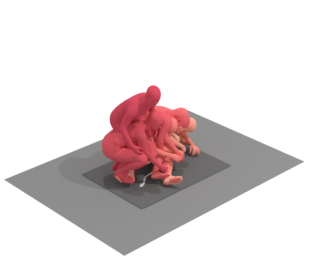}
 \textcolor{green}{\fbox{\includegraphics[width=0.092\linewidth]{figs/results/t2m_000284/000284.png}}}
 \includegraphics[width=0.092\linewidth]{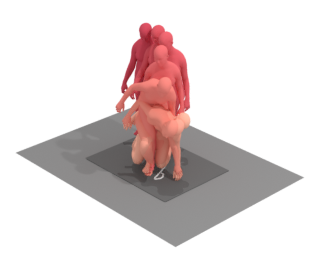}
 \includegraphics[width=0.092\linewidth]{figs/results/t2m_000284/003760.png}
 \includegraphics[width=0.092\linewidth]{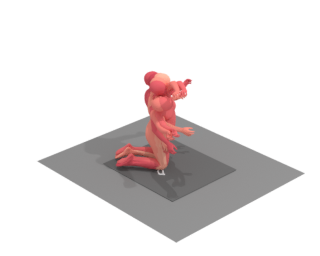}
 \includegraphics[width=0.092\linewidth]{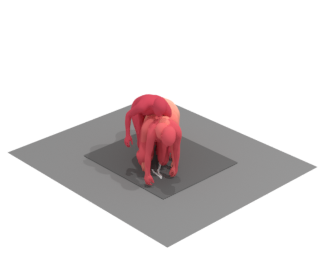}
 \includegraphics[width=0.092\linewidth]{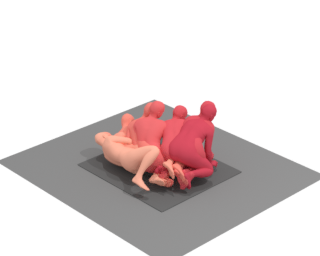}
 \includegraphics[width=0.092\linewidth]{figs/results/t2m_000284/001616.png} & \raisebox{2.5ex}{\color[HTML]{2D79A3}$\heartsuit$} \\ \hline 

  \multirow{2}{0.23\textwidth}{\textit{a person walks forward while twisting their torso side to side.}} & \includegraphics[width=0.092\linewidth]{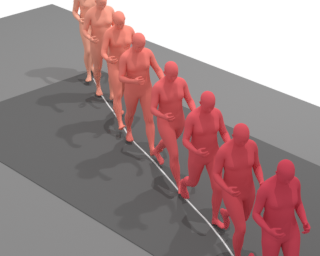}
\includegraphics[width=0.092\linewidth]{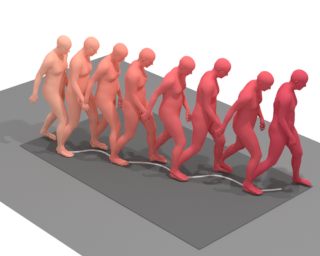}
\includegraphics[width=0.092\linewidth]{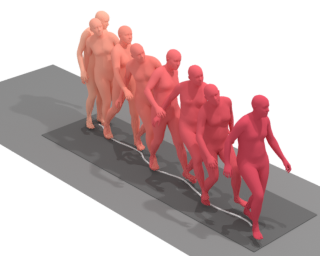}
\includegraphics[width=0.092\linewidth]{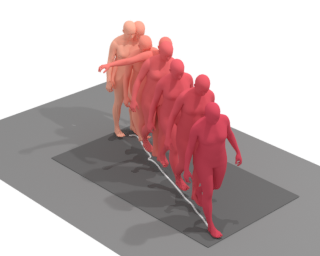}
\includegraphics[width=0.092\linewidth]{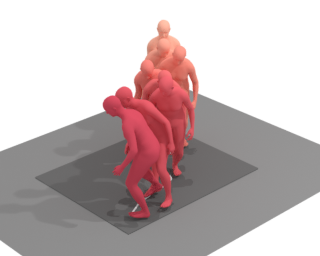}
\includegraphics[width=0.092\linewidth]{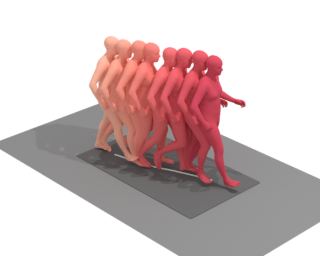}
\includegraphics[width=0.092\linewidth]{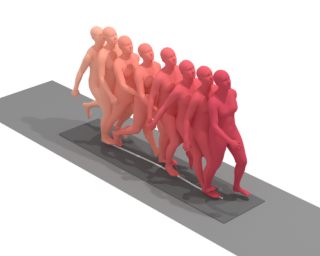}
\includegraphics[width=0.092\linewidth]{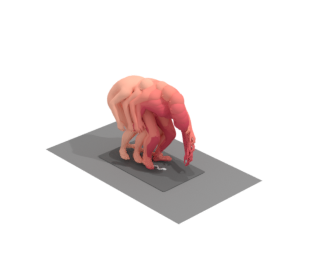}
\includegraphics[width=0.092\linewidth]{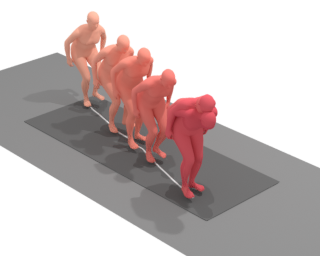}
\textcolor{green}{\fbox{\includegraphics[width=0.092\linewidth]{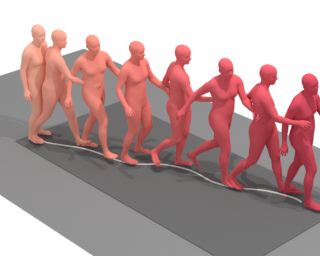}}} &  \raisebox{2.5ex}{\color[HTML]{C8414B}$\diamondsuit$} \\
 & \textcolor{green}{\fbox{\includegraphics[width=0.092\linewidth]{figs/results/t2m_000710/000710.png}}}
 \includegraphics[width=0.092\linewidth]{figs/results/t2m_000710/009584.png}
 \includegraphics[width=0.092\linewidth]{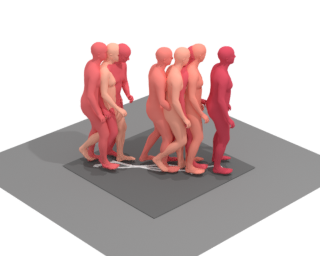}
 \includegraphics[width=0.092\linewidth]{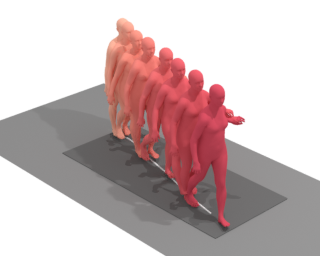} 
 \includegraphics[width=0.092\linewidth]{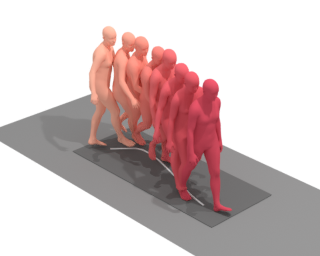}
 \includegraphics[width=0.092\linewidth]{figs/results/t2m_000710/010127.png}
 \includegraphics[width=0.092\linewidth]{figs/results/t2m_000710/005366.png}
 \includegraphics[width=0.092\linewidth]{figs/results/t2m_000710/003856.png}
 \includegraphics[width=0.092\linewidth]{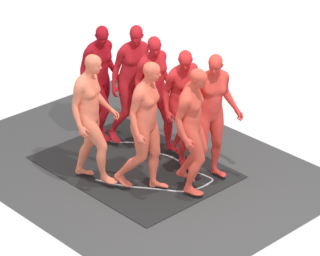}
 \includegraphics[width=0.092\linewidth]{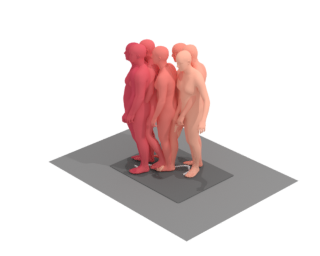} & \raisebox{2.5ex}{\color[HTML]{2D79A3}$\heartsuit$} \\ \hline

 \multirow{2}{0.23\textwidth}{\vspace{2mm}\\ \textit{a person turns to the left and gets down on his hands and crawls forward, towards the left, then crawls back to the area he started and gets up.}} & \vspace{1mm} \includegraphics[width=0.092\linewidth]{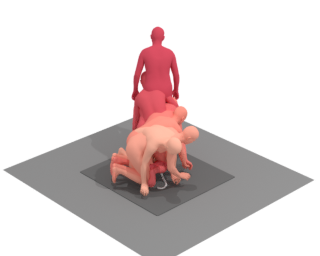}
\includegraphics[width=0.092\linewidth]{figs/results/t2m_000199/013589.png}
\includegraphics[width=0.092\linewidth]{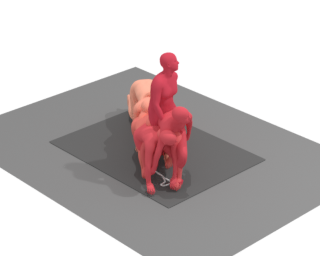}
\includegraphics[width=0.092\linewidth]{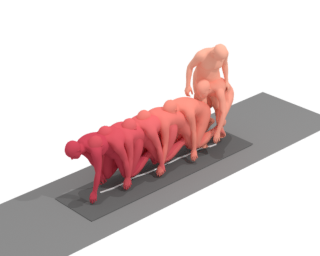}
\includegraphics[width=0.092\linewidth]{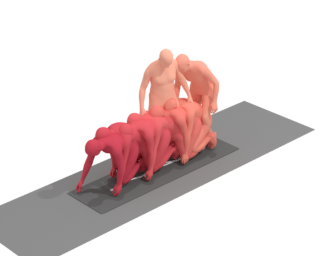}
\includegraphics[width=0.092\linewidth]{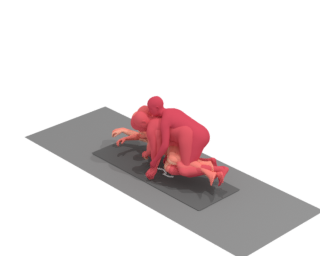}
\includegraphics[width=0.092\linewidth]{figs/results/t2m_000199/003142.png}
\textcolor{green}{\fbox{\includegraphics[width=0.092\linewidth]{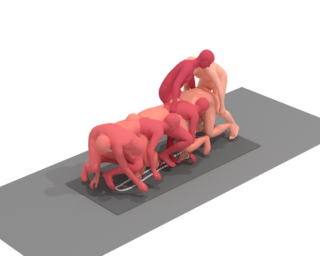}}}
\includegraphics[width=0.092\linewidth]{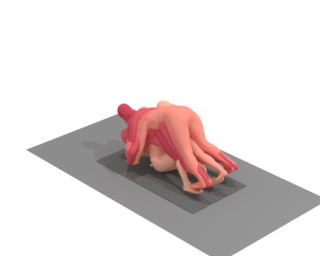}
\includegraphics[width=0.092\linewidth]{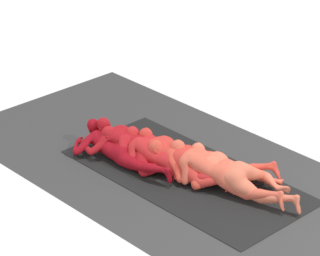} & \vspace{2mm} 
 \raisebox{-5ex}{\color[HTML]{C8414B}$\diamondsuit$} \\
 & \textcolor{green}{\fbox{\includegraphics[width=0.092\linewidth]{figs/results/t2m_000199/000199.png}}}
 \includegraphics[width=0.092\linewidth]{figs/results/t2m_000199/008275.png}
 \includegraphics[width=0.092\linewidth]{figs/results/t2m_000199/003142.png}
 \includegraphics[width=0.092\linewidth]{figs/results/t2m_000199/013589.png}
 \includegraphics[width=0.092\linewidth]{figs/results/t2m_000199/009662.png}
 \includegraphics[width=0.092\linewidth]{figs/results/t2m_000199/013662.png}
 \includegraphics[width=0.092\linewidth]{figs/results/t2m_000199/004007.png}
 \includegraphics[width=0.092\linewidth]{figs/results/t2m_000199/002294.png}
 \includegraphics[width=0.092\linewidth]{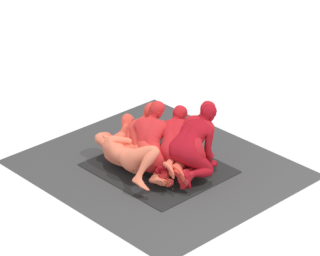}
 \includegraphics[width=0.092\linewidth]{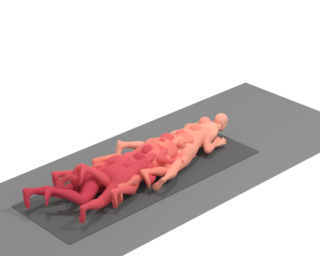} 
 & \raisebox{2.5ex}{\color[HTML]{2D79A3}$\heartsuit$}  \\ \bottomrule
 \multicolumn{3}{l}{\begin{tabular}[c]{@{}l@{}}\color[HTML]{C8414B}$\diamondsuit$ SH Loss \\ \color[HTML]{2D79A3}$\heartsuit$ Our DropTriple Loss \end{tabular}}
\end{tabular}%
\vspace{-5pt}
\end{table*}

\section{Experiments} \label{sec:experiments}

In this section, we introduce the standard datasets and evaluation protocols (Section \ref{ssec:dataset}), the representation of human poses used in our experiments (Section \ref{ssec:pose_representation}), experimental implementation details (Section \ref{ssec:implement_details}), and the experimental results (Section \ref{ssec:results}), including comparisons of loss functions, ablation studies, and more.

\subsection{Datasets, and Evaluation Protocol} \label{ssec:dataset}

We conducted experimental evaluations on two datasets: HumanML3D \cite{guo2022generating} and KIT-ML \cite{plappert2016kit}.

\textbf{HumanML3D} is a recent dataset that was re-annotated in text form from the AMASS \cite{mahmood2019amass} and HumanAct12 \cite{guo2020action2motion} collections of motion capture data. It contains 14,616 motions annotated with 44,970 textual descriptions, with an average of 3 relevant descriptions per motion. As some descriptions only cover parts of certain motions due to their complexity, we consider these sub-motions and corresponding textual descriptions as additional ground-truth pairs. Thus, the dataset contains 15,541 motions. We used the downsampled data to 20 fps and split the dataset into a training set with 14,541 motions and a test set with 1,000 motions.

\textbf{KIT Motion-Language (KIT-ML)} is composed of subsets of the KIT \cite{mandery2015kit} and the CMU datasets. It contains 3,911 motions and 6,353 sequence-level descriptions, with an average of 9.5 words per description. Among them, 3,008 sequences are valid with textual annotations, and each motion has 1-8 relevant descriptions, totaling 6,349 textual descriptions. We used the downsampled data to 12.5 fps and split the dataset into a training set with 2,508 motions and a test set with 500 motions.

\textbf{Evaluation Metrics.} We evaluated the learned embeddings for cross-modal retrieval tasks based on several metrics, including Recall@K (R@K), Median Rank (Med R), and R-sum \cite{hodosh2013framing,gong2021limitations}. Given a query, we retrieve the K=[1, 5, 10] nearest neighbors from the database. Retrieval is considered successful if the correct sample is among the K nearest neighbors. R-sum is defined as follows:
\begin{equation} 
\mathrm{\text{R-sum}}=\overbrace{\mathrm{R}@1+\mathrm{R}@5+\mathrm{R}@10}^{\text {Motion Retrieval }}+\overbrace{\mathrm{R}@1+\mathrm{R}@5+\mathrm{R}@10}^{\text {Text Retrieval }} 
\end{equation}


\subsection{Pose Representation} \label{ssec:pose_representation} 

We adopt the redundant pose representation provided in \cite{guo2022generating}. A pose $f$ is defined by the tuple $({{\dot{r}}^{a}},{{\dot{r}}^{x}},{{\dot{r}}^{z}},{{\dot{r}}^{y}},{{\mathbf{j}}^{p}},{{\mathbf{j}}^{v}},{{\mathbf{j}}^{r}},{{\mathbf{c}}^{f}})$, where ${{\dot{r}}^{a}}\in \mathbb{R}$ is the root angular velocity along the Y-axis; $({{\dot{r}}^{x}},{{\dot{r}}^{z}})\in \mathbb{R}$ are the root linear velocities in the XZ plane; ${{\dot{r}}^{y}}\in \mathbb{R}$ is the root height; ${{\mathbf{j}}^{p}}\in {{\mathbb{R}}^{3j}}$, ${{\mathbf{j}}^{v}}\in {{\mathbb{R}}^{3j}}$ and ${{\mathbf{j}}^{r}}\in {{\mathbb{R}}^{6j}}$ are the local joint rotation-invariant position \cite{holden2016deep}, velocity, and 6D continuous rotation \cite{zhou2019continuity} in the root space, where $j$ is the number of joints; and ${{\mathbf{c}}^{f}}\in {{\mathbb{R}}^{4}}$ is a binary feature obtained by thresholding the velocities of the heel and toe joints to emphasize foot-ground contact. The motion of the HumanML3D dataset follows the skeleton structure of 22 joints in SMPL \cite{loper2015smpl}, and each motion sequence is represented as $\boldsymbol{m} \in \mathbb{R}^{F \times 263}$. For the KIT-ML dataset, the poses have 21 joints, and $\boldsymbol{m} \in \mathbb{R}^{F \times 251}$. All human motion sequences are initially facing the Z+ direction.

\begin{figure}[t]
    \centering
    \vspace{-10pt}
    \begin{minipage}[t]{0.32\linewidth}
        \centering
        \includegraphics[width=0.7\linewidth]{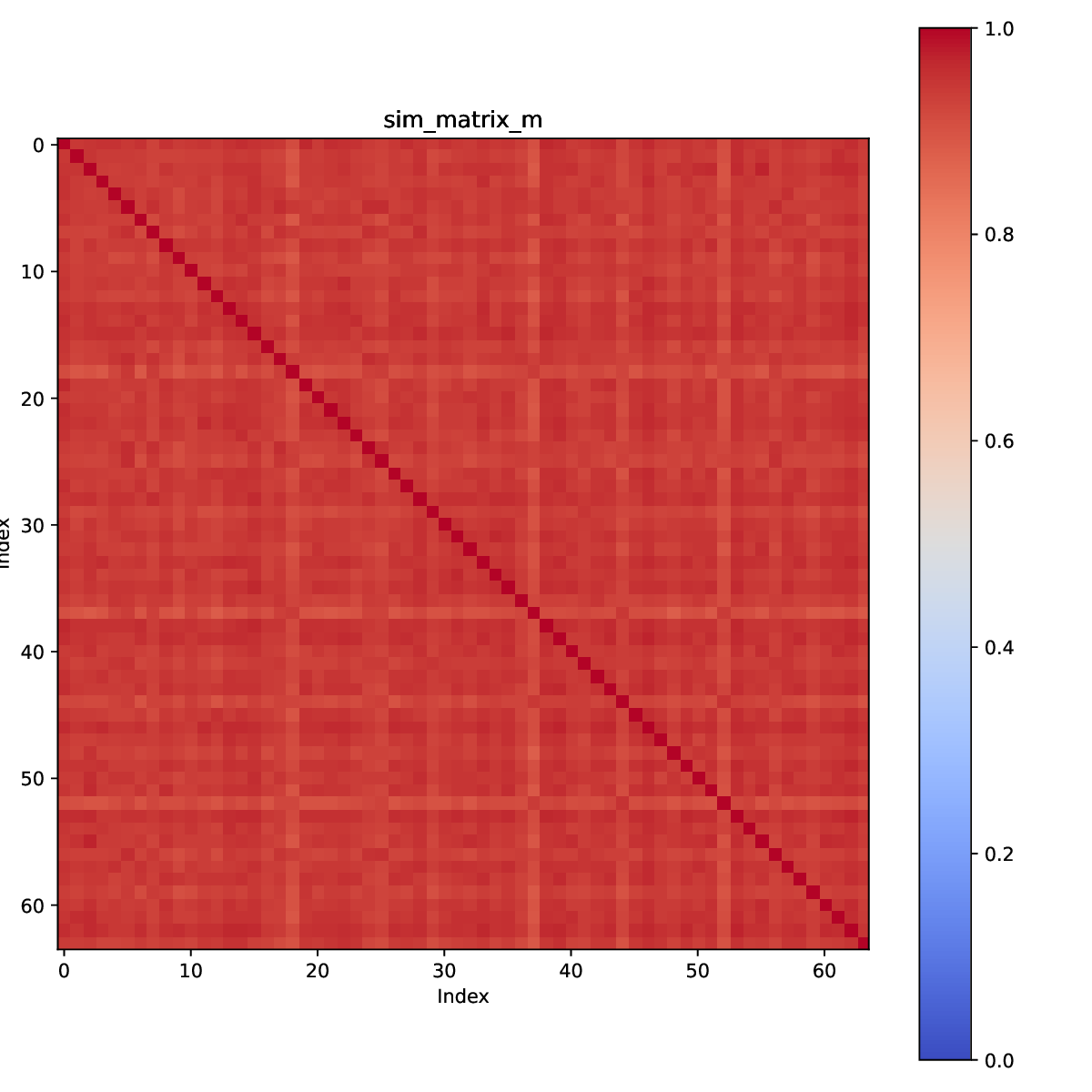}
        \vspace{-5pt}
        \subcaption{epoch-1}
        \label{fig:inter_m_epoch1}
    \end{minipage}
    \hfill
    \begin{minipage}[t]{0.32\linewidth}
        \centering
        \includegraphics[width=0.7\linewidth]{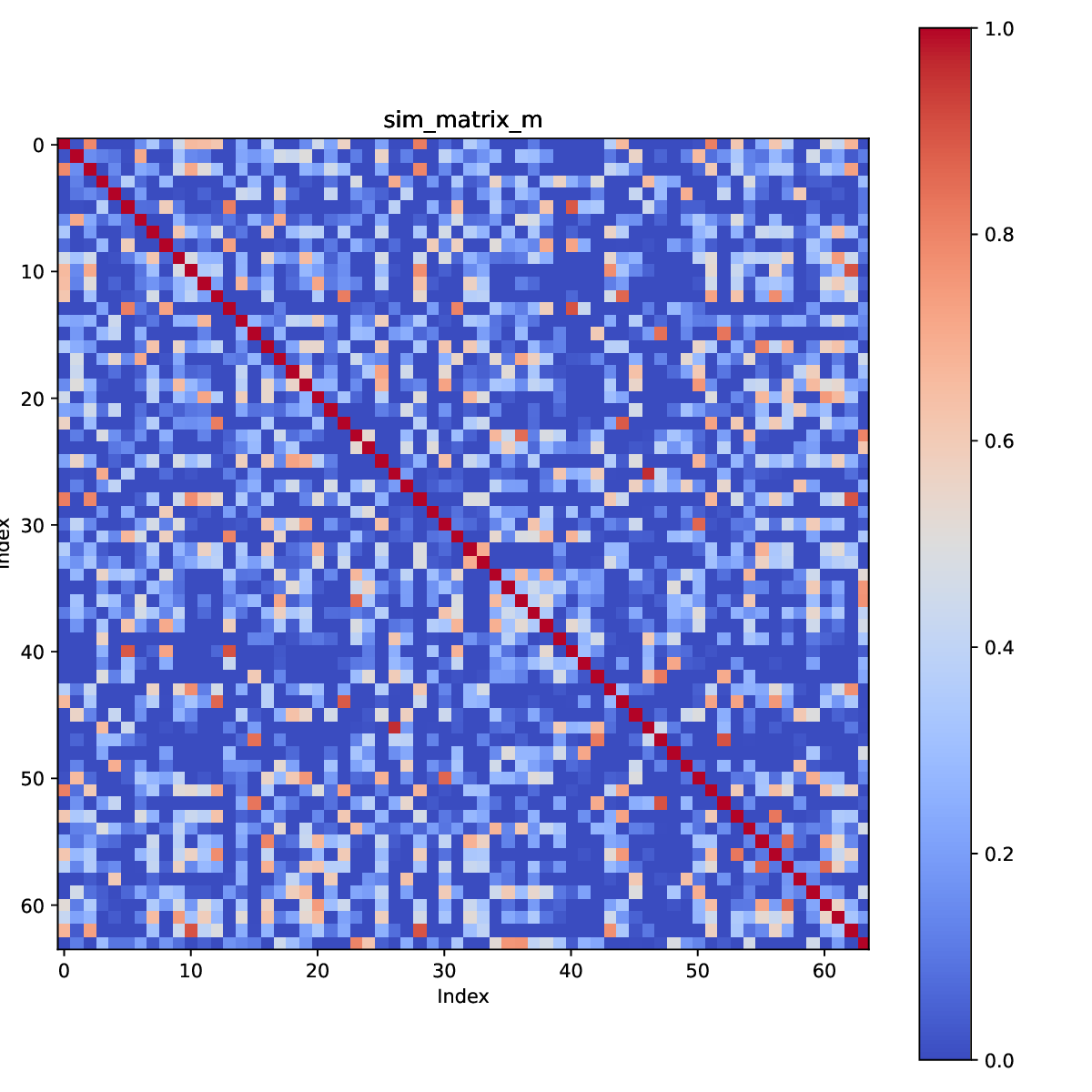}
        \vspace{-5pt}
        \subcaption{epoch-6}
        \label{fig:inter_m_epoch5}
    \end{minipage}
    \hfill
    \begin{minipage}[t]{0.32\linewidth}
        \centering
        \includegraphics[width=0.7\linewidth]{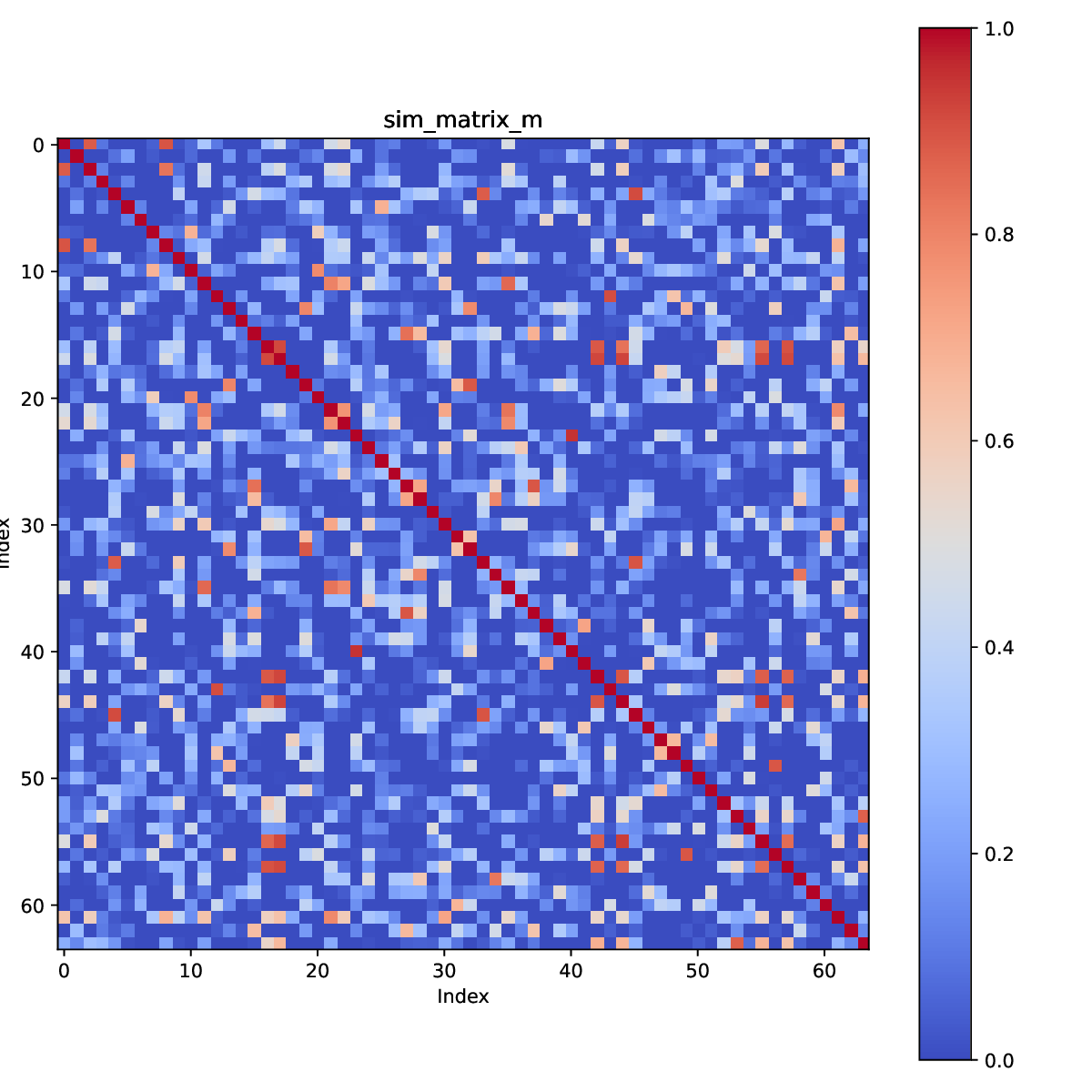}
        \vspace{-5pt}
        \subcaption{epoch-11}
        \label{fig:inter_m_epoch10}
    \end{minipage}
    \vskip\baselineskip
    \vspace{-10pt}
    \begin{minipage}[t]{0.32\linewidth}
        \centering
        \includegraphics[width=0.7\linewidth]{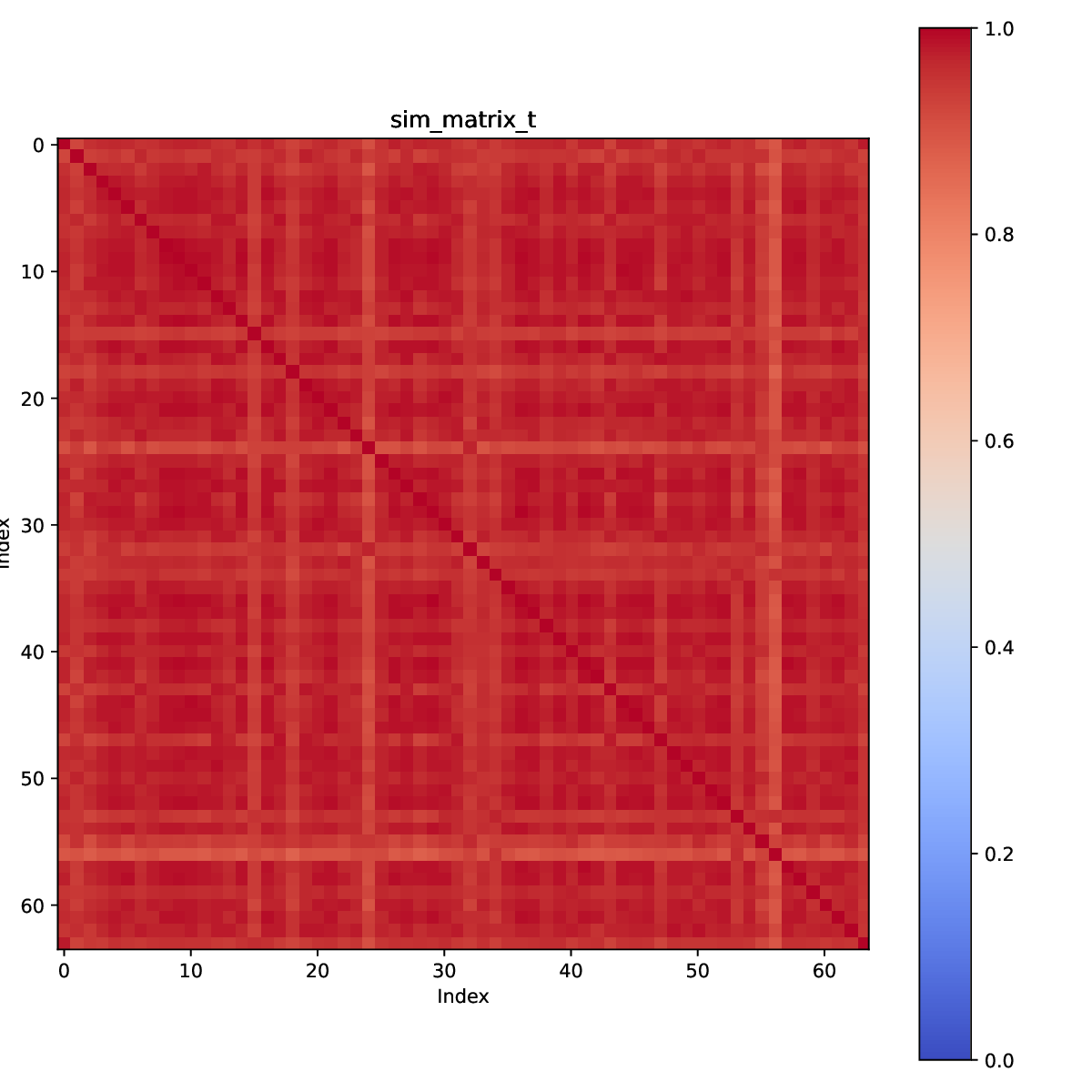}
        \vspace{-5pt}
        \subcaption{epoch-1}
        \label{fig:inter_t_epoch1}
    \end{minipage}
    \hfill
    \begin{minipage}[t]{0.32\linewidth}
        \centering
        \includegraphics[width=0.7\linewidth]{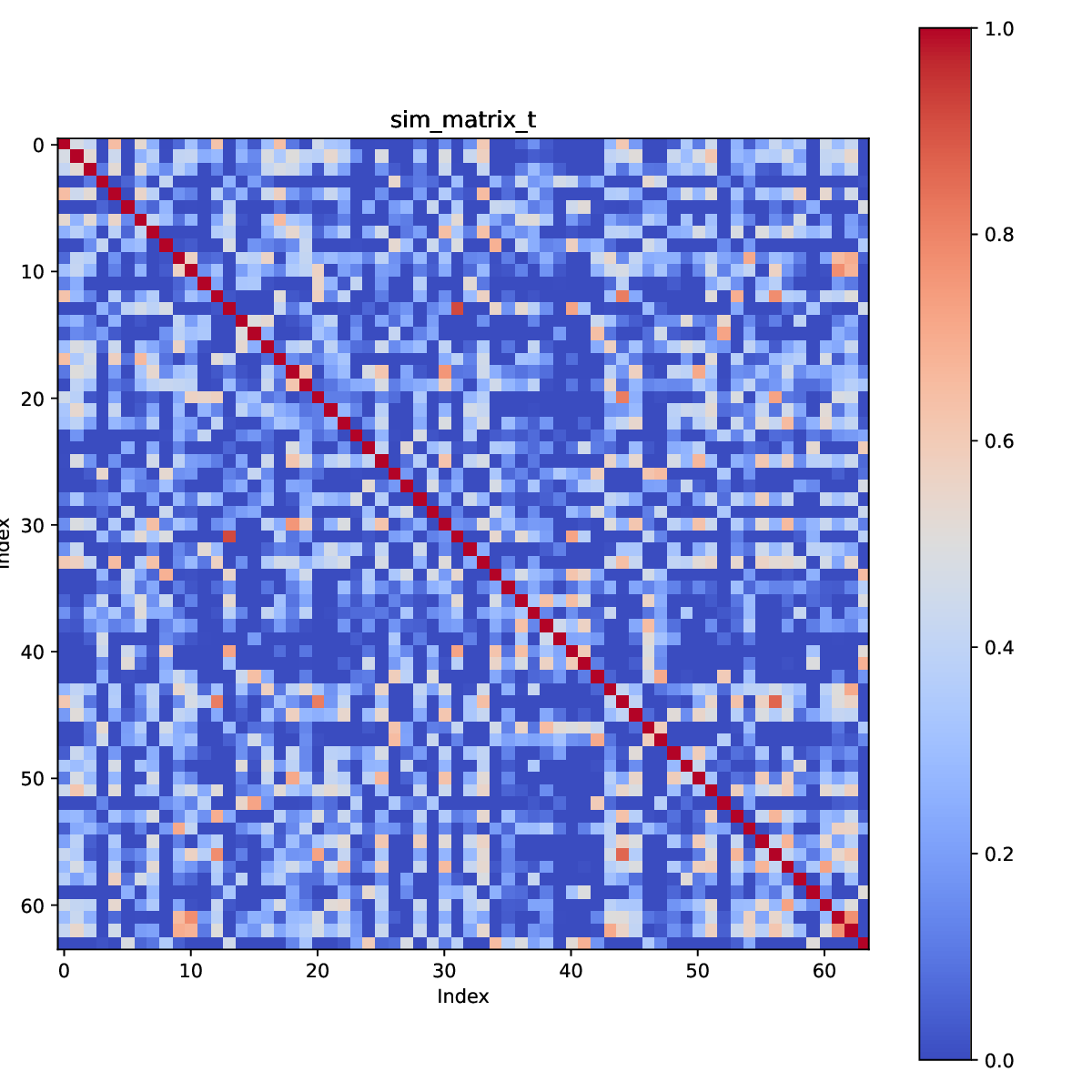}
        \vspace{-5pt}
        \subcaption{epoch-6}
        \label{fig:inter_t_epoch5}
    \end{minipage}
    \begin{minipage}[t]{0.32\linewidth}
        \centering
        \includegraphics[width=0.7\linewidth]{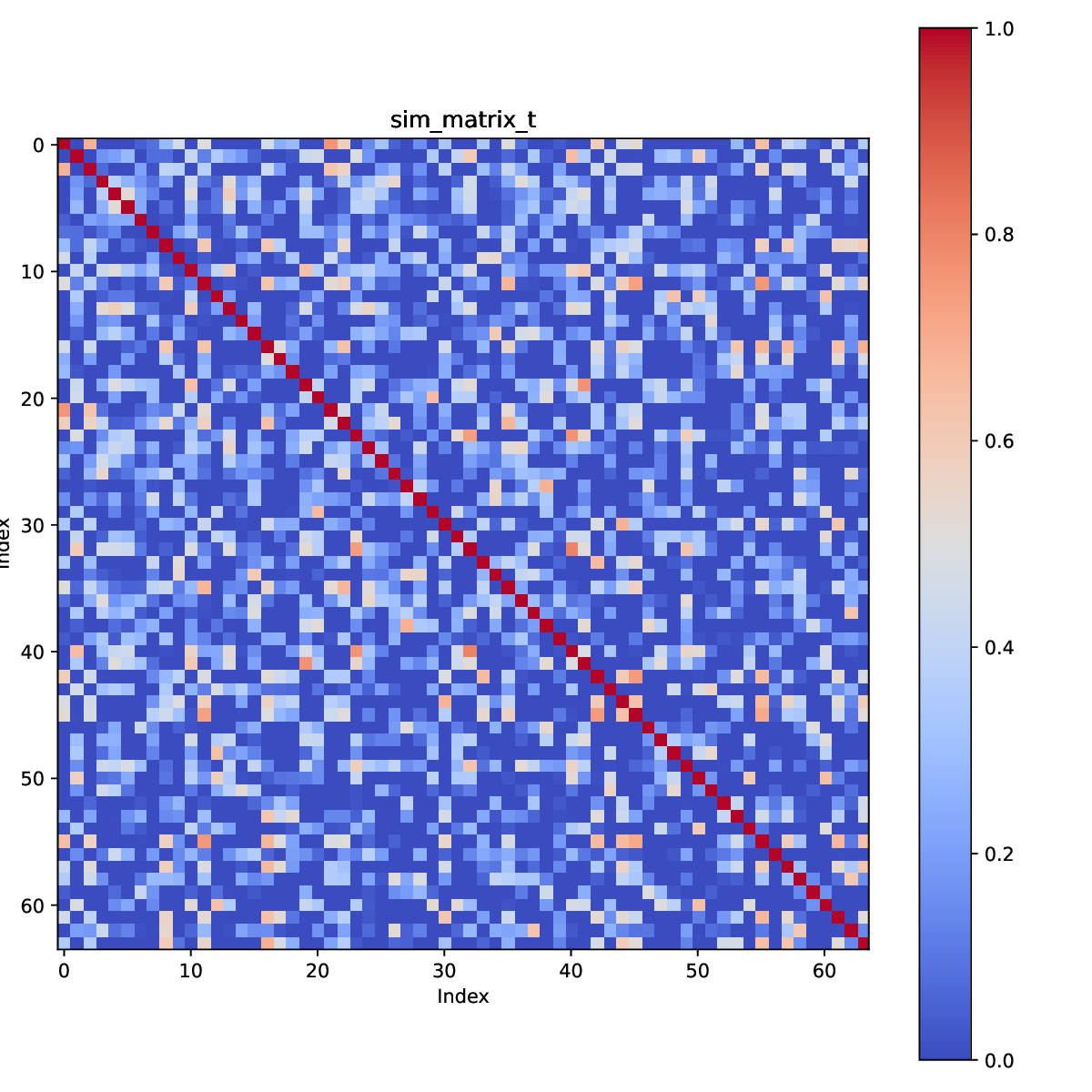}
        \vspace{-5pt}
        \subcaption{epoch-11}
        \label{fig:inter_t_epoch10}
    \end{minipage}
    \vspace{-10pt}
    \caption{Intra-Modal similarity matrices of a batch in different training epochs on HumanML3D. (a), (b), and (c): motion modality, (d), (e), and (f): text modality.}
    \label{fig:similar_matrix}
    \vspace{-10pt}
\end{figure}

\begin{table*}[h]
\centering
\caption{Text retrieval results on HumanML3D. We visualize top 5 results of a given query. The ground-truth is highlighted in blue.}
\label{tab:text_retrieval}
\vspace{-10pt}
\resizebox{\textwidth}{!}{%
\begin{tabular}{c p{0.4\textwidth} p{0.4\textwidth}}
\toprule
\textbf{Query} & \multicolumn{2}{c}{\textbf{Top-5 Retrieved Texts}} \\ \hline
\raisebox{-0.5\height}{\includegraphics[width=0.2\linewidth]{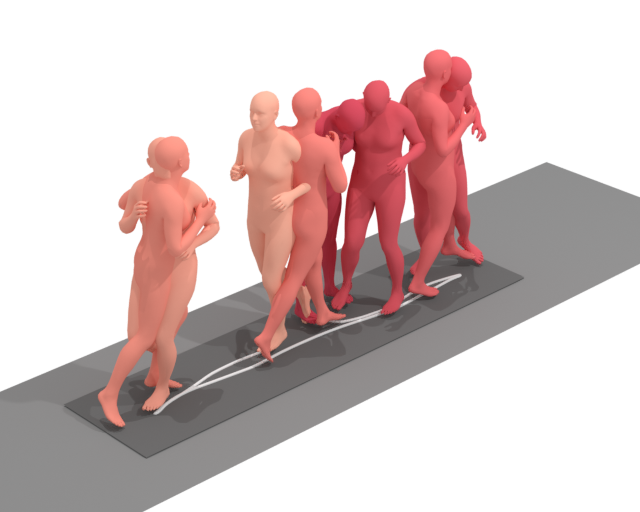}} & \multicolumn{1}{l}{\begin{tabular}[c]{@{}>{\raggedright}p{0.4\textwidth}@{}}
1. the person is jogging back-and-forth to the left and right.\\ 
2. a person jogs back and forth. \\ 
3. a person jogs back and forth.\\ 
\textcolor{blue}{4. person is jogging from left to right and then back to the center.}\\ 
5. a person ran in left and after right direction and returned.\end{tabular}} & \begin{tabular}[c]{@{}>{\raggedright}p{0.4\textwidth}@{}}
1. the person is jogging back-and-forth to the left and right.\\ 
\textcolor{blue}{2. a person starts in the middle, runs to the right, then to the left, then returns back to their starting position.}\\ 
\textcolor{blue}{3. person is jogging from left to right and then back to the center.}\\ 
4. a person jogs back and forth.\\ 
5. a person jogs back and forth.\end{tabular} \\ \hline

\raisebox{-0.5\height}{\includegraphics[width=0.2\linewidth]{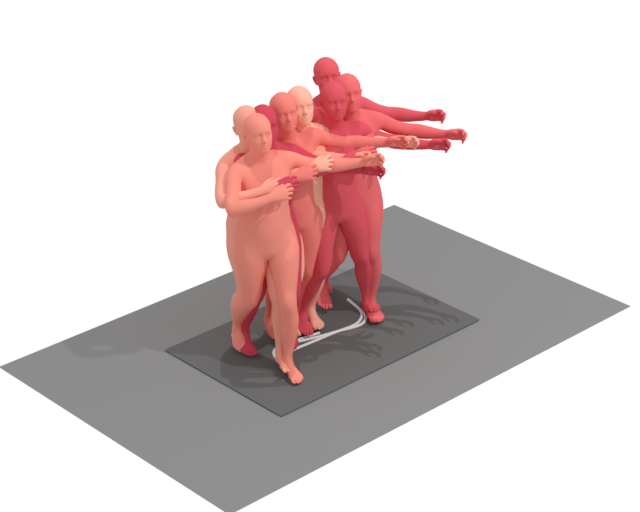}} & \multicolumn{1}{l}{\begin{tabular}[c]{@{}>{\raggedright}p{0.4\textwidth}@{}}
1. walking while swinging both arms from left to right.\\ 
2. a person is dancing with left hand holding someone.\\ 
3. holding a partner, a person dances a waltz.\\ 
4. a person strums a guitar/banjo with their right hand while holding the neck in their left.\\ 
\textcolor{blue}{5. a person is doing the cha cha dance.}\end{tabular}} & \begin{tabular}[c]{@{}>{\raggedright}p{0.4\textwidth}@{}}
\textcolor{blue}{1. a person performs the waltz.}\\ 
2. holding a partner, a person dances a waltz.\\ 
3. a person is dancing with left hand holding someone.\\ 
4. a figure waltzes, taking large steps in a circle rhythmically.\\ 
5. a person is dancing the cha cha.\end{tabular} \\ \hline

\raisebox{-0.5\height}{\includegraphics[width=0.2\linewidth]{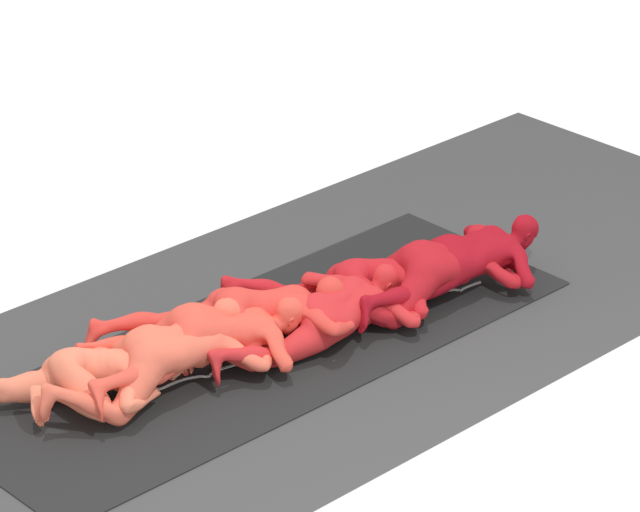}} & \multicolumn{1}{l}{\begin{tabular}[c]{@{}>{\raggedright}p{0.4\textwidth}@{}}
1. the person is crawling in their hands and knees.\\ 
2. a person drops down to their hands and knees and proceeds to crawl forward.\\ 
\textcolor{blue}{3. a man crawls forward on his stomach.}\\ 
4. laying down on face and crawling backwards.\\ 
5. a man gets on his hands and knees and crawls forward.\end{tabular}} & \begin{tabular}[c]{@{}>{\raggedright}p{0.4\textwidth}@{}}
\textcolor{blue}{1. a man crawls forward on his stomach.}\\ 
\textcolor{blue}{2. this person crawls on all fours with belly close to the ground.}\\ 
3. a person crawls along the ground on their belly.\\ 
4. the person is crawling in their hands and knees.\\ 
5. laying down on face and crawling backwards.\end{tabular} \\ \bottomrule

 & \multicolumn{1}{c}{\color[HTML]{C8414B} SH Loss} & \multicolumn{1}{c}{\color[HTML]{2D79A3} Our DropTriple Loss}
\end{tabular}%
}
\vspace{-10pt}
\end{table*}

\subsection{Implementation Details} \label{ssec:implement_details}

We used a learning rate of 2e-4 to train for 60 epochs with AdamW optimizer \cite{kingma2014adam}, including a warm-up of $\rho=5$ epochs. The learning rate was reduced by a factor of 10 at the 30th epoch. Due to limited computational resources, we excluded motion sequences exceeding 1000 frames from training and testing (downsampled). The joint embedding space $D$ was set to 1024 with a margin $\alpha$ of 0.2. For the motion branch, the backbone network dimension $D_{l}$ was set to 256 to align with the embedding layer dimension. TMR Enc had 1 layer for KIT-ML and 3 layers for HumanML3D. For the text branch, we used DistilBert \cite{sanh2019distilbert} to generate 768-dimensional features ($D_{w}=768$). The batch size was set to 32 for KIT-ML and 64 for HumanML3D. Empirically, we set $\delta _{hetero}$ and $\delta _{homo}$ to 0.6 and 0.9 for KIT-ML, 0.7 and 0.9 for HumanML3D. To address the slow convergence issue when training the model with MH Loss on HumanML3D, we extended the learning rate decay to the 45th epoch for fair comparison and continued training for an additional 30 epochs. During the fine-tuning of the DistilBERT language model parameters, we used a learning rate of 2e-4 for its network and trained for an additional 30 epochs.

\begin{figure}[t]
    \vspace{-15pt}
    \centering
    \begin{subfigure}[t]{0.48\linewidth}
        \centering
        \includegraphics[width=\linewidth]{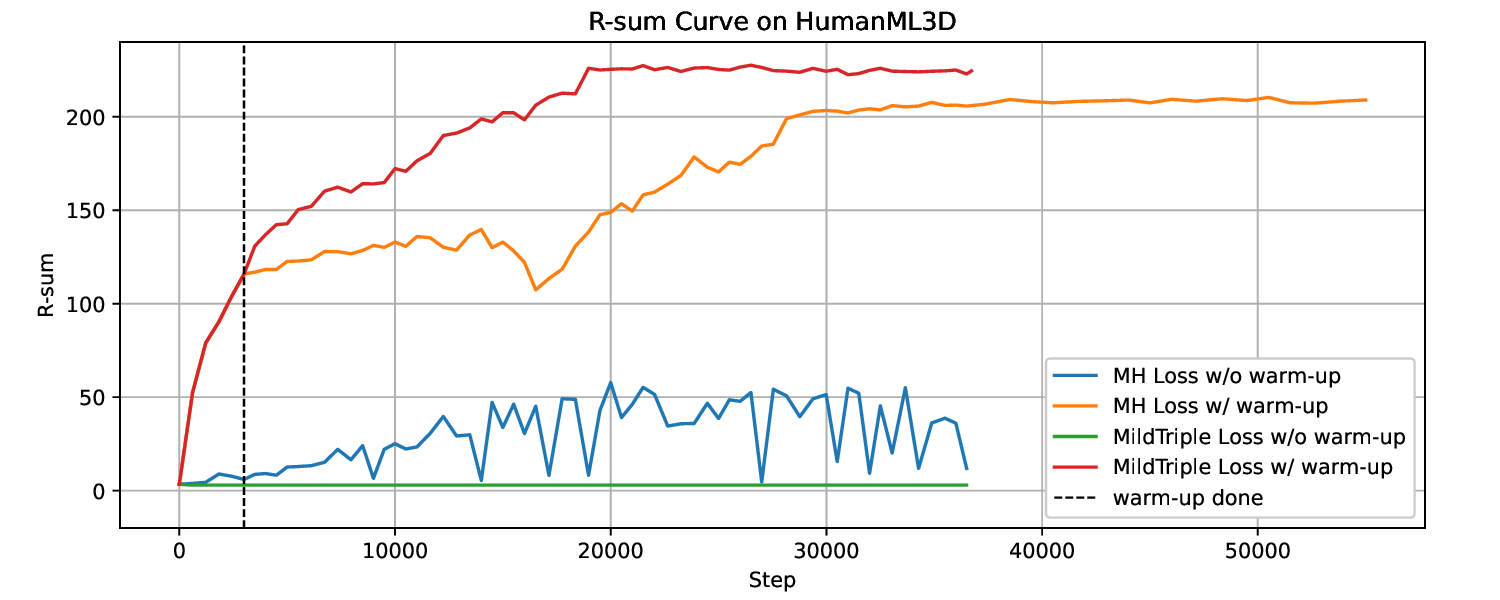}
        \caption{R-sum}
        \label{fig:warmup_loss}
    \end{subfigure}
    \hfill
    \begin{subfigure}[t]{0.48\linewidth}
        \centering
        \includegraphics[width=\linewidth]{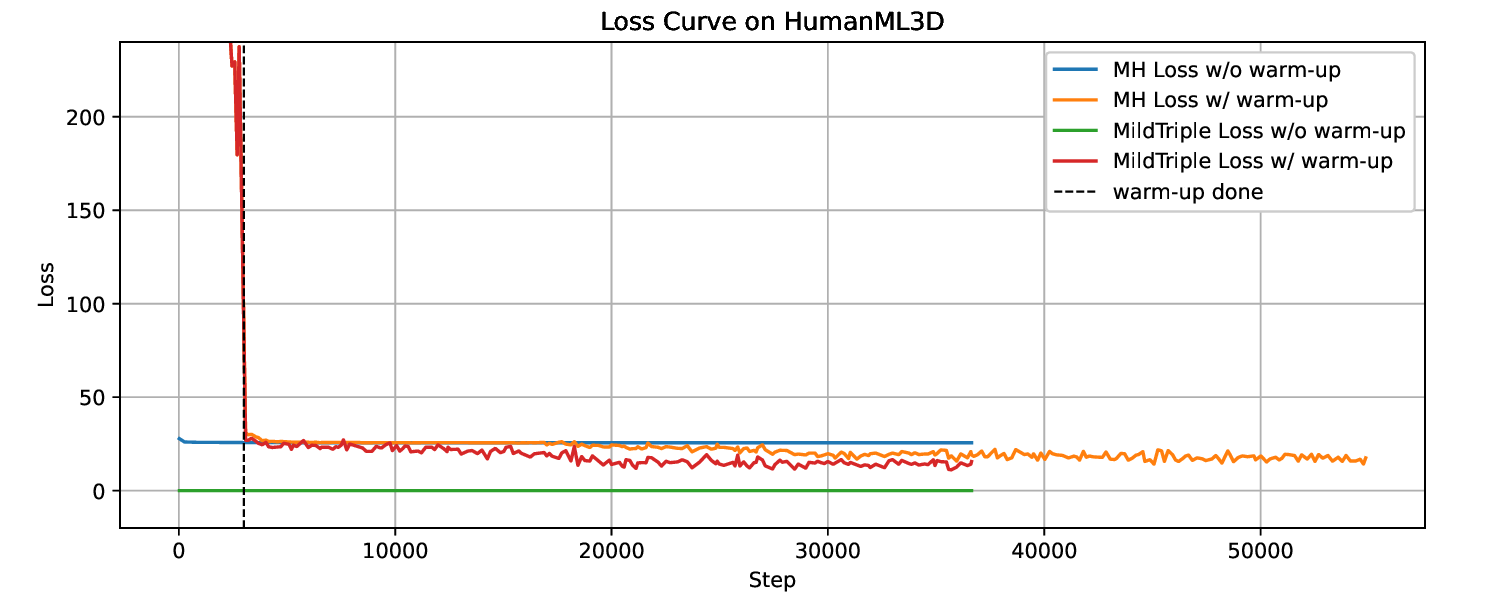}
        \caption{Loss curve}
        \label{fig:warmup_rsum}
    \end{subfigure}
    \vspace{-10pt}
    \caption{R-sum and loss variation when using MH Loss or DropTriple Loss w/ and w/o warm-up on HumanML3D.}
    \label{fig:warmup}
    \vspace{-5pt}
\end{figure}

\begin{figure}[t]
    \centering
    \vspace{-10pt}
    \begin{subfigure}[t]{0.48\linewidth}
        \centering
        \includegraphics[width=0.8\linewidth]{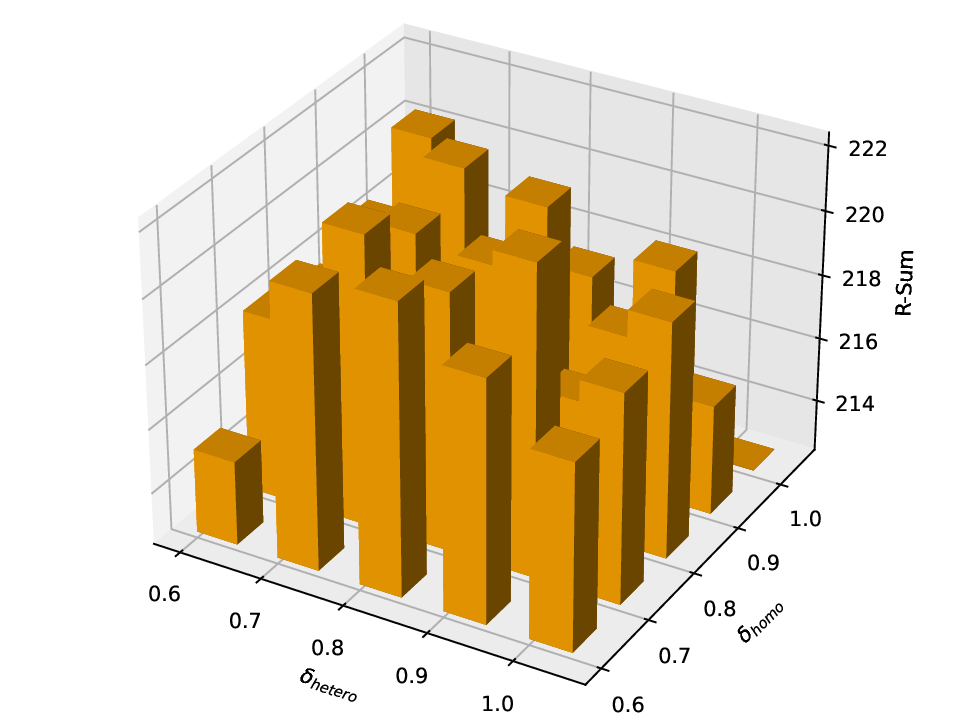}
        \caption{HumanML3D}
        \label{fig:rsum_h3d}
    \end{subfigure}
    \hfill
    \begin{subfigure}[t]{0.48\linewidth}
        \centering
        \includegraphics[width=0.8\linewidth]{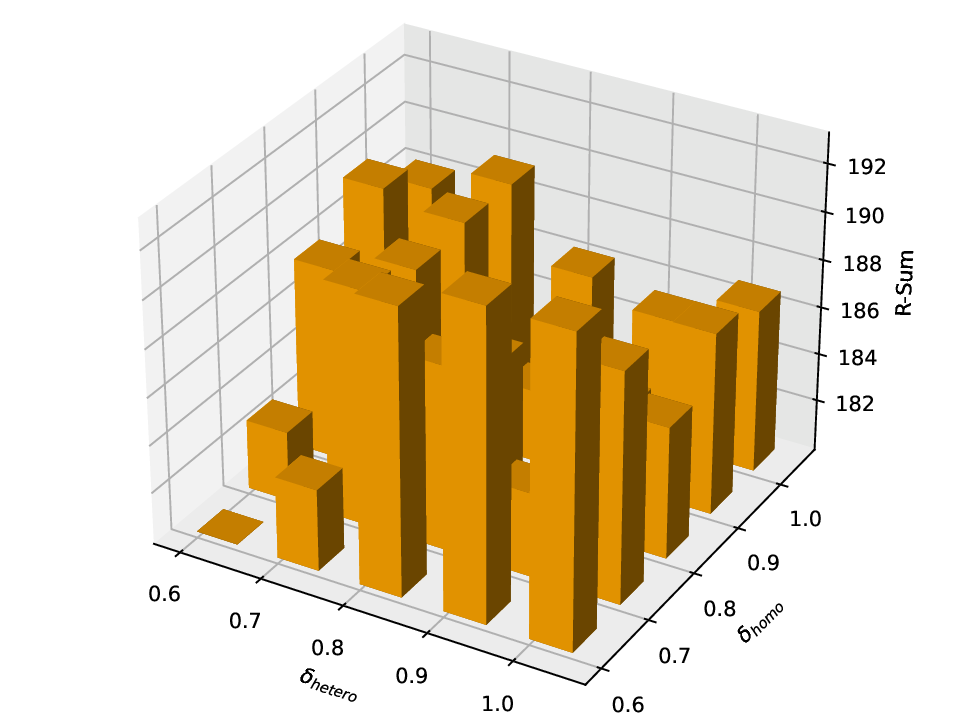}
        \caption{KIT-ML}
        \label{fig:rsum_kit}
    \end{subfigure}
    \vspace{-10pt}
    \caption{R-sum results obtained using different thresholds on the two datasets.}
    \label{fig:rsum}
    \vspace{-22pt}
\end{figure}

\subsection{Results} \label{ssec:results}

We present the results of motion-text bidirectional retrieval using SH Loss, MH Loss, and DropTriple Loss in Table \ref{tab:reasult}. It can be observed from the tables that DropTriple Loss consistently outperforms SH Loss and MH Loss on both datasets. On the HumanML3D dataset, DropTriple Loss achieves improvements of 5.3\%, 9.9\%, and 8.8\% in terms of R@1, R@5, and R@10 (sum of motion and text retrieval results), respectively, compared to SH Loss. Additionally, the Med R is reduced by 2.0. Comparing the experimental data, we observe that while MH Loss slightly improves the performance of the motion-text retrieval task, it fails to achieve the expected results due to excessive optimization of false-Negs, particularly resulting in a decrease in R-sum compared to SH Loss on KIT-ML. In contrast, our DropTriple Loss mitigates the adverse effects of optimizing false-Negs, thereby yielding more effective results.

Finally, fine-tuning the language model based on the DropTriple Loss further enhances the model's performance. All metrics of the fine-tuned DropTriple Loss method achieve their highest values, particularly on the HumanML3D dataset, where R@5 and R@10 for motion retrieval increase by 10.0\%, and R-sum reaches 276.4.

\subsubsection{Ablation Study on Warm-up and Threshold $\delta$} \label{sec:ablation_study}

We investigated the importance of warm-up , as depicted in Figure \ref{fig:warmup}, which illustrates the R-sum scores and loss variations when using MH Loss or DropTriple Loss w/ and w/o warm-up. After warm-up, the R-sum score steadily increases and the loss decreases.  Training with MH Loss w/o warm-up was slow because it relies on a smaller set of triplets compared to SH Loss. Early in training, the gradient of MH Loss was influenced by a relatively small set of triplets, requiring more iterations to train the model with MH Loss. For the case where the loss remained zero when using DropTriple Loss w/o warm-up, we visualize the similarity matrices of each modality at different training epochs (see Figure \ref{fig:similar_matrix}). In the early epoch of training (epoch 1), both matrices (Figures \ref{fig:inter_m_epoch1}, \ref{fig:inter_t_epoch1} ) are mostly red, indicating that the similarity between samples exceeds the threshold $\delta$ by we set, resulting in the inclusion of all Negs in the pruned set of false-Negs. As a result, there is no optimization target available.

We varied the thresholds $\delta_{homo}$ and $\delta_{hetero}$ to assess their impact on DropTriple Loss. Figure \ref{fig:rsum} presents the R-sum results obtained using different thresholds on the two datasets. By observing the changes in R-sum, it can be seen that when both $\delta_{homo}$ and $\delta_{hetero}$ are small, the results are similar to or even lower than those obtained by training the model with MH Loss (i.e., $\delta_{homo}=1.0$ and $\delta_{hetero}=1.0$). This suggests that when the thresholds are below a certain level, the inclusion of some ordinary Negs in the false-Neg set can lead to a decrease in DropTriple Loss' ability to select hard-Negs.

\subsubsection{Qualitative Results}

To demonstrate the effectiveness of our model and DropTriple Loss, we conducted a qualitative comparison between the previous SH Loss and DropTriple Loss. As shown in Tables \ref{tab:motion_retrieval} and \ref{tab:text_retrieval}, our model performs well in retrieving ground-truth samples. Furthermore, the use of DropTriple Loss further improves the rankings. Specifically, comparing the first row results in Table \ref{tab:text_retrieval}, DropTriple Loss focuses on optimizing the genuinely hard-Negs: \textit{"a person jogs back and forth."}, placing them below the ground-truth.

\section{Conclusion} \label{sec:conclusion}

In this work, we make a meaningful attempt to investigate the overlooked task of motion-text cross-modal retrieval by constructing a concise yet effective model. Additionally, we proposed DropTriple Loss and validated its ability to reduce the semantic conflicts caused by false negative samples in triplet training. In future work, we aim to explore the potential of extending the DropTriple Loss to other domain retrieval tasks.

\begin{acks}
This work was supported by the National Natural Science Foundation of China (No. 62203476).
\end{acks}

\bibliographystyle{ACM-Reference-Format}
\bibliography{sample-base}

\setcounter{table}{0}

\end{document}